\pdfoutput=1

\documentclass[11pt]{article}

\usepackage[final]{acl}

\usepackage{times}
\usepackage{latexsym}

\usepackage[T1]{fontenc}

\usepackage[utf8]{inputenc}

\usepackage{microtype}

\usepackage{inconsolata}

\usepackage{amsmath,bm}
\usepackage{amssymb}
\usepackage{mathrsfs}
\usepackage{algpseudocode}
\usepackage{graphicx}
\usepackage{epstopdf}
\usepackage{adjustbox}
\usepackage{enumitem}
\usepackage{latexsym}
\usepackage{subfigure}
\usepackage{xcolor}
\usepackage{booktabs}

\usepackage{soul}
\usepackage{color, xcolor}
\title{ItD: Large Language Models Can Teach Themselves \\ Induction through Deduction}

\author{
  Wangtao Sun$^{1,2}$, Haotian Xu$^{6}$, Xuanqing Yu$^{2,3}$, Pei Chen$^{4}$, Shizhu He$^{1,2}$, Jun Zhao$^{1,2}$, Kang Liu$^{1,2,5}$ \\
  \textit{$^{1}$The Laboratory of Cognition and Decision Intelligence for Complex Systems,} \\
  \textit{Institute of Automation, Chinese Academy of Sciences, Beijing, China} \\
  \textit{$^{2}$School of Artificial Intelligence, University of Chinese Academy of Sciences, Beijing, China} \\
  \textit{$^{3}$CAS Engineering Laboratory for Intelligent Industrial Vision,} \\
  \textit{Institute of Automation, Chinese Academy of Sciences, Beijing, China} \\
  \textit{$^{4}$Department of Computer Science and Engineering, Texas A\&M University} \\
  \textit{$^{5}$Shanghai Artificial Intelligence Laboratory} \\
  \textit{$^{6}$Xiaohongshu Inc} \\
  \texttt{sunwangtao2021@ia.ac.cn} \\
}

\begin{document}

\maketitle

\begin{abstract}
Although Large Language Models (LLMs) are showing impressive performance on a wide range of Natural Language Processing tasks, researchers have found that they still have limited ability to conduct induction. Recent works mainly adopt ``post processes'' paradigms to improve the performance of LLMs on induction (e.g., the hypothesis search \& refinement methods), but their performance is still constrained by the inherent inductive capability of the LLMs. In this paper, we propose a novel framework, \underline{I}nduction \underline{t}hrough \underline{D}eduction (ItD), to enable the LLMs to teach themselves induction through deduction. 
The ItD framework is composed of two main components: a Deductive Data Generation module to generate induction data and a Naive Bayesian Induction module to optimize the fine-tuning and decoding of LLMs. 
Our empirical results showcase the effectiveness of ItD on two induction benchmarks, achieving relative performance improvement of 36\% and 10\% compared with previous state-of-the-art, respectively. Our ablation study verifies the effectiveness of two key modules of ItD. We also verify the effectiveness of ItD across different LLMs and deductors. The data and code of this paper can be found at https://anonymous.4open.science/r/ItD-E844.
\end{abstract}



\section{Introduction}
\label{intro}
\begin{figure}[t]
    \centering
    \begin{adjustbox}{max width=\columnwidth}    
    \includegraphics[width=0.45\textwidth]{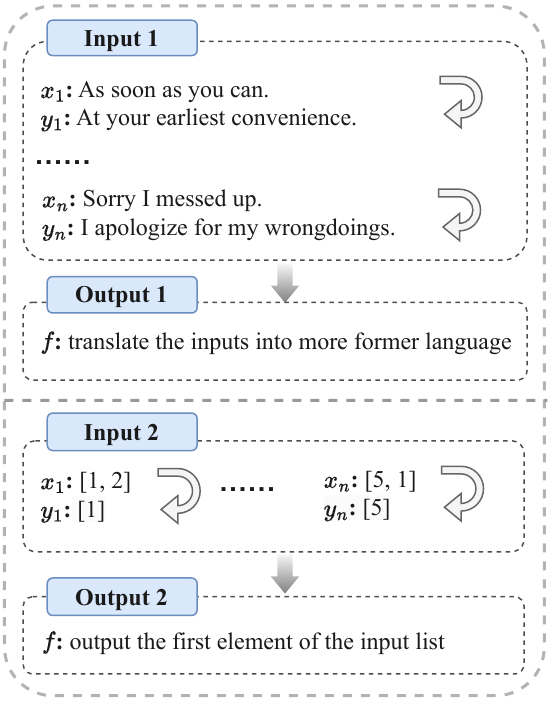}
    \end{adjustbox}
    \caption{Task of Induction. The tested model observes a batch of input-output $(x,y)$ pairs and needs to predict the latent transformation $f$ shared by these $(x,y)$ pairs.}
    \label{task}
\end{figure}

Induction can take we humans from the observed to the unobserved \cite{sloman2005problem}. The task of \textbf{\textit{Induction}} aims to discover consistent transformations from a set of input-output pairs, where the transformations map the inputs to the outputs well \cite{wang2023hypothesis}. As shown in Figure~\ref{task},
given the input-output pairs $\{x_i,y_i\}_{i=1}^n$, the model needs to predict the latent transformation $f$. For a detailed example,
given the input \emph{[1,2]} with the output \emph{[1]} and other input-output pairs, the tested model is supposed to figure out the transformation \emph{output the first element of the input list}. The Induction task is an important task in Natural Language Processing (NLP) and the mastery of the induction ability is an important sign of intelligence \cite{peirce1868questions,lake2017building, chollet2019measure}.

Currently, humans have already mastered the capability of induction and have found thousands of laws from the physical world and human society. 
However, machine intelligence still struggles to induce basic logic rules in structure data like knowledge graphs \cite{nskg, grzymala2023rule}. Recently, with the rapid development of Large Language Models (LLMs), many works have begun to adopt the LLMs to induce the transformations given the input-output observations of various tasks and express the induced transformations as rules \cite{yang2023failures, sun2023expnote, hit, zhao2023expel}, guidelines \cite{pang2023guideline}, instructions \cite{honovich2022instruction}, and codes \cite{progres, wang2023hypothesis}. These methods take advantage of the interpretability and generalization ability of LLMs in solving the Induction task.





\begin{figure}[t]
    \centering
    \begin{adjustbox}{max width=\columnwidth}    
    \includegraphics[width=0.5\textwidth]{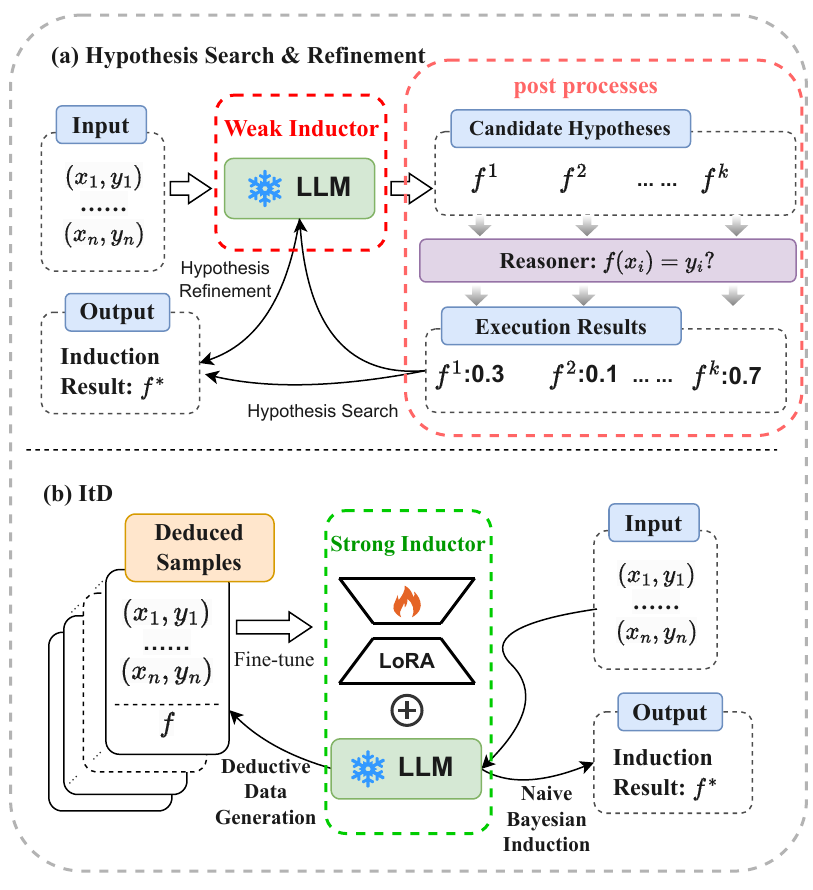}
    \end{adjustbox}
    \caption{Comparison of ItD and Previous Methods. Previous hypothesis search \& refinement methods are essentially ``post processes" to the raw induction results of LLMs, leaving LLMs as Weak Inductors. By contrast, ItD fine-tune the LLMs and propose a novel decoding algorithm to make them Strong Inductors.}
    \label{comparison}
\end{figure}

However, recent research~\cite{bang2023multitask, mitchell2023comparing, eval1, eval2} have revealed that LLMs have inherently limited ability in induction. To tackle such a limitation, work like Hypothesis Search \cite{wang2023hypothesis} proposes to select the generated hypotheses from LLMs by evaluating them on the observations, while another following work Iterative Hypothesis Refinement \cite{qiu2023phenomenal} proposes to further refine them through LLMs based on the evaluating results on the observations. Nevertheless, as shown in Figure~\ref{comparison}(a), these hypothesis search \& refinement methods are essentially ``post processes'' to the directly induced hypotheses of LLMs. They still heavily rely on the inherent induction ability of LLMs which are Weak Inductors.



Even though LLMs are limited in induction, recent work finds out that they possess much better capability in deduction~\cite{bang2023multitask, semanticThanSymbolic}. 
Different from induction, deduction aims to infer the correct output given the transformation and the input.
Despite the distinction that induction associates multiple $(x, y)$ pairs with the latent transformation $f$, whereas deduction links $x$ and $f$ to the resultant $y$, both approaches fundamentally share the commonality of reasoning within the framework of input, output, and transformation $(x, y, f)$.
Therefore, it motivates us to propose a novel framework ItD (\underline{I}nduction \underline{t}hrough \underline{D}eduction), to enable the LLMs to teach themselves induction through deduction. 
Different from previous methods, ItD fine-tunes the LLMs on their deduced data to make them Strong Inductors, as shown in Figure~\ref{comparison}(b). 
For a given induction task, ItD first proposes \emph{Deductive Data Generation} 
to leverage the deductive capability of the LLMs to generate a set of task data $(x,y,f)$, 
which is simple yet effective and does not rely on human annotations or any larger LLMs' assistance. 
The data will then be used to fine-tune the LLMs to obtain better inductive capability.





However, it is non-trivial to utilize the deduced data. We find out that directly fine-tuning the LLMs using the IO prompt used in the previous methods~\cite{honovich2022instruction, wang2023hypothesis, qiu2023phenomenal} cannot effectively leverage the observed samples (as shown in Figure~\ref{sample_number}). Thus, ItD further proposes \emph{Naive Bayesian Induction} as a strategy to optimize the use of each sample. Moreover, we also observe performance gains with the increase in the number of samples using our approach.
Specifically, this novel technique fine-tunes the LLM to predict $f$ conditioned on single pair $x,y$ ($p(f|x,y)$) instead of $n$ pairs ($p(f|\{x_i,y_i\}_{i=1}^n)$). During the decoding phase, it utilizes the Naive Bayesian approach to equivalently infer the probability distribution of $f$ under all $n$ $(x,y)$ conditions ($p(f|\{x_i,y_i\}_{i=1}^n)$) with the probability distribution of $f$ under a single $(x,y)$ condition ($p(f|x,y)$).

We conduct experiments on two different types of induction tasks for evaluation: Instruction Induction and List Function.
Compared with previous methods, The experiment results show that ItD is superior to the existing methods in assisting LLMs in induction, and both the Deductive Data Generation and the Naive Bayesian Induction components effectively contribute to ItD. 
We also make discussions to show that ItD can be effectively applied to different LLMs, and a more powerful deductor, e.g. \texttt{ChatGPT}, will further improve the performances of ItD. In summary, the major contributions of this paper are as follows:

\begin{figure*}[t]
    \centering
    \begin{adjustbox}{max width=\textwidth}    
    \includegraphics[width=1.05\textwidth]{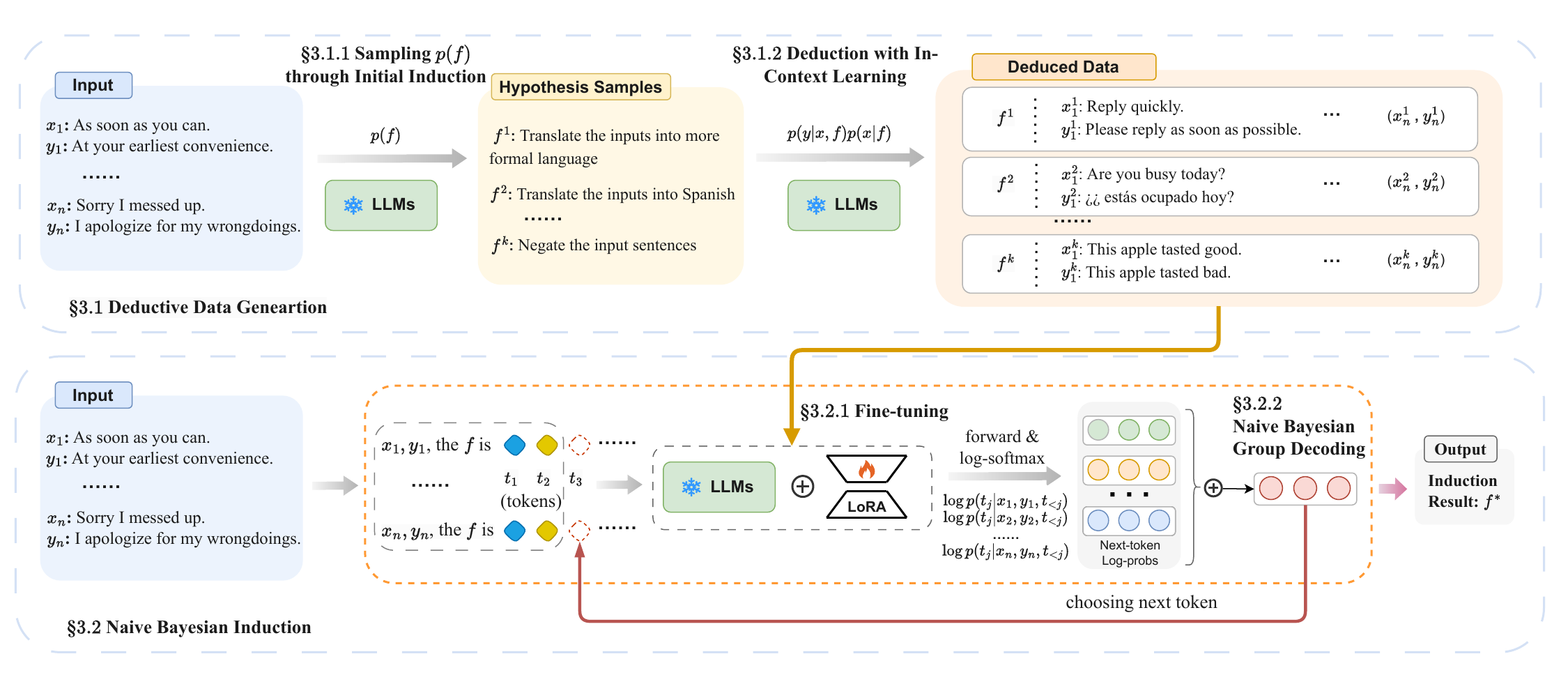}
    \end{adjustbox}
    \caption{The framework of ItD. ItD includes two main parts, i.e. Deductive Data Generation and Naive Bayesian Induction. Given the induction set $\mathcal{D}_{in}$, ItD will first leverage the deductive capability of LLMs to generate data that closely resembles the distribution of the task data. Then Naive Bayesian Induction is adopted to optimize the use of each observed sample while achieving better performances with the increase in the number of samples.}
    \label{framework}
\end{figure*}

\begin{itemize}[itemsep=1pt,topsep=1pt,parsep=0pt,leftmargin=*]
    \item We propose a novel framework ItD to enable the LLMs to teach themselves induction through deduction.
    \item We propose Deductive Data Generation to effectively leverage the deductive capability of LLMs to generate task data. which is fully self-supervised and needs no human annotations or any larger LLMs to assist. 
    \item We propose Naive Bayesian Induction to allow LLMs to optimize the use of each observed sample and be able to take advantage of the increase in the number of observed samples.
\end{itemize}

\section{Preliminary}

\subsection{Induction Task}
As shown in Figure~\ref{task}, induction aims to infer the latent transformation, $f$, from a few of observed samples, $\{x_i,y_i\}_{i=1}^n$, where $y_i=f(x_i)$.

An induction task $\mathcal{T}$ will include multiple input-output data pairs $\mathcal{D} = \{x_{i},y_{i}\}_{i=1}^m$, and all $(x_{i},y_{i})$ share the same latent ground truth transformation $f$. The complete task data $\mathcal{D}$ of task $\mathcal{T}$ is then split into an induction set $\mathcal{D}_{in}$ and a deduction set $\mathcal{D}_{de}$.

The testing model is asked to first run the induction process on $\mathcal{D}_{in}$. $\mathcal{D}_{in}$ is split into multiple batches, with each batch containing $n$ samples $\{x_i,y_i\}_{i=1}^n$. The batches will be fed into the model sequentially. The testing model observes the input batches and induces the predicted transformation $f^*$. All $f^*$ induced from $\mathcal{D}_{in}$ will be collected for deduction.

In the deduction process, a shared Reasoner $\mathcal{R}$ is used to execute all induced $f^*$ from different methods on $\mathcal{D}_{de}$ for fairness. For all test samples $(x_{test}, y_{test})$ from $\mathcal{D}_{de}$, the candidate $f^*$ and test input $x_{test}$ will be fed into $\mathcal{R}$ and then $\mathcal{R}$ generates the prediction $y^*$. Finally, we evaluate the metric between $y_{test}$ and $y^*$ and average it over $f^*$.

\subsection{IO Prompt}
\label{io prompt}

As the induction task offers the model $n$ observed samples at a time, it is natural to organize the samples into the IO (Input-Output) prompt as follows: $x_1, y_1; x_2, y_2; ...; x_n, y_n$, which is also widely used by previous works \cite{honovich2022instruction, wang2023hypothesis, qiu2023phenomenal}. Note that we omit the instructions and other connection words in the prompt above. For example, for the \emph{Input2} in Figure~\ref{task}, the IO prompt can be: \emph{Please figure out the transformation that transforms the following input lists to the output lists: Input:[1,2], Output:[1], ......, Input:[5,1], Output:[5]. So the transformation is:}.


\begin{figure}[t]
    \centering
    \includegraphics[width=0.5\textwidth]{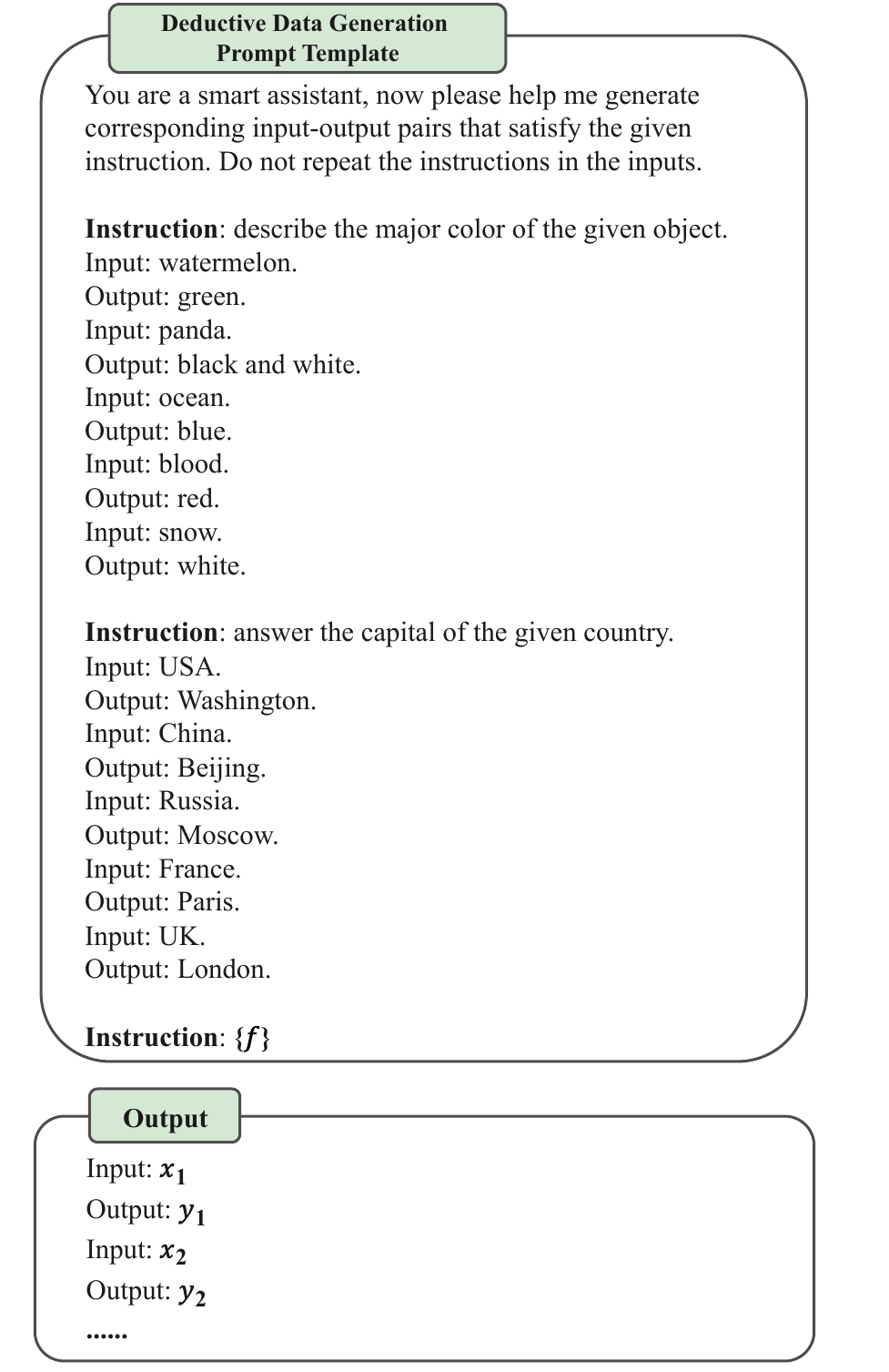}
    \caption{The prompt used for Deduction with In-Context Learning. LLMs will generate multiple samples $(x,y)$ for each $f$ in a deductive behavior.}
    \label{icl}
\end{figure}

\section{Framework}

In this section, we introduce ItD, a framework for empowering the induction capability of LLMs. As shown in Figure~\ref{framework}, ItD is composed of two modules: Deductive Data Generation, and Naive Bayesian Induction. For a given induction task, Deductive Data Generation will first leverage the deductive capability of the LLMs to generate the task data. Then we propose Naive Bayesian Induction to allow LLMs to optimize the use of each observed sample, while taking advantage of the increase in the number of observed samples.

\subsection{Deductive Data Generation}
\label{data generation}


To empower the induction ability of LLMs, a set of training data $(x,y,f)$ is needed. Here we consider sampling from their joint distribution $p(x,y,f)$.
As we introduced in \S\ref{intro}, compared with induction $p(f|x,y)$, LLMs are better at deduction $p(y|f,x)$. Thus we propose the following derivation to leverage the LLMs to generate the task data in a deductive behavior.

\begin{equation}
\label{eq1}
    \begin{aligned}
        p(x,y,f) &= p(x,y|f)p(f) \\
                 &= p(y|x,f)p(x|f)p(f)
    \end{aligned}
\end{equation}


As shown in Eq~(\ref{eq1}), to generate data $(x,y,f)$, we propose to sample $p(f)$, $p(x|f)$, and $p(y|x,f)$ sequentially. The pipeline of Deductive Data Generation is shown in the upper half of Figure~\ref{framework}.

\subsubsection{Sampling $p(f)$ through Initial Induction}
\label{prior}


To ensure that the generated data $(x,y,f)$ approximates the real task data distribution, we first need to sample the transformation $f$ that approximates the ground truth transformation of the task. 
Thus, we let LLMs induce $f$ on the induction set $\mathcal{D}_{in}$ in the sampling decoding mode with the IO prompt, and we consider the induced $f$ as samples from the prior distribution $p(f)$.

\subsubsection{Deduction with In-Context Learning}
\label{DICL}

For the $p(y|x,f)p(x|f)$ part in Eq~(\ref{eq1}), this paper leverages the deductive capability with In-Context Learning (ICL) of LLMs to generate samples $(x,y)$. We first manually create several cases of deduction as the few-shot demonstrations and then ask LLMs to generate samples $(x,y)$ for each $f$ (Figure~\ref{icl}).




As shown in Figure~\ref{icl}, the upper half is the fixed prompt and the content of the last instruction is replaced by each $f$ from \S\ref{prior}. 
In the lower half, the LLMs will follow the demonstrations to continuously first generate an input $x_i$ according to the instruction $f$, and then generate $y_i$ based on their deductive capability. We then parse the output text of LLMs to obtain the samples $\{x_i,y_i\}_{i=1}^n$. 
The deductive capabilities of LLMs will determine the extent to which $(x, y)$ satisfies the given $f$.
For each $f$, we generate $n$ corresponding $(x,y)$ pairs for later tuning.

\subsection{Naive Bayesian Induction}
\label{Naive Bayesian Induction}
Having obtained the generated task data, we propose Naive Bayesian Induction which incorporates tuning and decoding to empower the inductive capability of LLMs.
The pipeline of Naive Bayesian Induction is shown in the lower half of Figure~\ref{framework}.
Instead of the plain IO prompt (\S\ref{io prompt}), Naive Bayesian Induction proposes the Group Decoding (GD) prompt template as follows: $x, y$. Compared with the IO prompt, the GD prompt contains only one input-output pair $(x,y)$.

By using the GD prompt in Naive Bayesian Induction, we allow LLMs to optimize the use of each observed sample ($p(f|x,y)$) and can take advantage of the increase in the number of observed samples.
Naive Bayesian Induction further proposes Naive Bayesian Group Decoding, which enables us to equivalently infer the probability distribution of $f$ under all $n$ $(x,y)$ conditions ($p(f|\{x_i,y_i\}_{i=1}^n)$) with the probability distribution of $f$ under a single $(x,y)$ condition ($p(f|x,y)$). 

Specifically, the IO prompt and GD prompt fine-tune the LLM and decode with the LLMs according to the following distribution respectively.

\begin{itemize}[itemsep=1pt,topsep=1pt,parsep=0pt]
    \item \textbf{IO prompt}: $p_{LLM}(f|\{x_i,y_i\}_{i=1}^n)$
    \item \textbf{GD prompt}: $p_{LLM}(f|x,y)$
\end{itemize}

\subsubsection{Fine-tuning on the Deduced Data}
For the shared fine-tuning data collected in \S\ref{data generation}, we organize them into the training data with IO prompt and GD prompt, respectively. Then, we adapt LoRA \cite{lora} and QLoRA \cite{dettmers2023qlora} to fine-tune the original LLMs to gain a better capability of induction. 

\subsubsection{Naive Bayesian Group Decoding}
\label{NBGD}

For the model trained with IO prompt, in the induction stage, we directly convert the $n$ observed sample from $\mathcal{D}_{in}$ into the IO prompt, feed it into the model, and use beam search to decode the $f$. This method is denoted as ItD-IO.

For the model trained with GD prompt, ItD proposes the following Naive Bayesian Group Decoding (NBGD) algorithm. NBGD allows us to equivalently infer the probability distribution of $f$ under all $n$ $(x,y)$ conditions ($p(f|\{x_i,y_i\}_{i=1}^n)$) with the probability distribution of $f$ under a single $(x,y)$ condition ($p(f|x,y)$). 

\begin{equation}
\label{eq2}
    \begin{aligned}
        p(f|\{x_i,y_i\}_{i=1}^n) &= \dfrac{p(\{x_i,y_i\}_{i=1}^n|f)p(f)}{p(\{x_i,y_i\}_{i=1}^n)} \\
                                 &\propto p(\{x_i,y_i\}_{i=1}^n|f)p(f) \\
                                 &=p(f)\prod_{i=1}^n p(x_i,y_i|f) \\
                                 &=p(f)\prod_{i=1}^n \dfrac{p(f|x_i,y_i)p(x_i,y_i)}{p(f)} \\
                                 &\propto p(f)^{-(n-1)}\prod_{i=1}^n p(f|x_i,y_i)
    \end{aligned}
\end{equation}
Here we assume that given the transformation $f$, the input-output pairs $(x,y)$ are independent to each other, i.e. $p(\{x_i,y_i\}_{i=1}^n|f) = \prod_{i=1}^n p(x_i,y_i|f)$. This assumption is quite natural in the scene of induction, where each $y_i$ is only determined by $f$ and the corresponding $x_i$.


As shown in Eq~(\ref{eq2}), we derive the probability $p(f|\{x_i,y_i\}_{i=1}^n)$ into two parts, i.e. the prior term $p(f)^{-(n-1)}$ and the product term $\prod_{i=1}^n p(f|x_i,y_i)$ respectively. Suppose the text of $f$ is a sentence $t=[t_1,t_2,...,t_m]$. For the $\prod_{i=1}^n p(f|x_i,y_i)$, we modify the ordinary beam search decoding process as follows:

\begin{equation}
\label{eq3}
    \begin{aligned}
        \sum_{i=1}^n \log p(t|x_i,y_i) &= \sum_{i=1}^n \sum_{j=1}^m \log p(t_j|x_i,y_i,t_{<j}) \\
                                       &= \sum_{j=1}^m \sum_{i=1}^n \log p(t_j|x_i,y_i,t_{<j})
    \end{aligned}
\end{equation}
As shown in Eq~(\ref{eq3}) and Figure~\ref{framework}, in the induction stage, NBGD will first convert all samples $(x_i,y_i)$ into GD prompt, tokenize them and feed them into the LLMs in a batch. Then in each step of decoding ($j$), the LLMs receive the already decoded part of transformation $t_{<j}$, and every sample $(x_i,y_i)$ and generate the next-token scores (log-probabilities) for $t_j$. Then we will add up the next-token scores from all the samples ($i$). Like ordinary beam search, in each step $j$, we will maintain the top-k beams with the largest beam scores.

After NBGD decodes the top-k $f$, we finally re-rank them through the prior term $p(f)^{-(n-1)}$. In the log scale, we only need to calculate the log probabilities $\log p(f)$ with the same LLMs and add $-(n-1)\log p(f)$ to their beam scores. We consider this training \& decoding method as the complete method of our framework, denoted as ItD.

\section{Experiments}



\subsection{Dataset and Setups}

We use two datasets to test the inductive capability of LLMs on two types of induction tasks: commonsense inductive reasoning and symbolic inductive reasoning.

For commonsense inductive reasoning, we adapt the task Instruction Induction \cite{honovich2022instruction}. The input $x$ and output $y$ are two short sentences while the transformation $f$ is an instruction. This dataset contains 24 sub-tasks.
For symbolic inductive reasoning, we adapt the task List Function \cite{listfunc}. The input $x$ and output $y$ are two integer lists while the transformation $f$ is a natural language description of the latent list transformation. This dataset contains 250 sub-tasks.

We adopt ChatGPT as the Reasoner $\mathcal{R}$ for both datasets and all tested methods. The reported results are average execution scores \cite{honovich2022instruction} over all sub-tasks. The detailed setups of the experiments can be found in Appendix~\ref{setups} and the detailed results of each sub-task of all methods can be found in Appendix~\ref{detailed results}.

\begin{table}[t]
\centering
\begin{adjustbox}{width=\columnwidth}
\begin{tabular}{lccc}
\toprule
Dataset  & Instruction Induction   & List Function          \\
\midrule
Model & Llama-2-7b-chat  & Mixtral-8x7B \\
\midrule
IO  &  13.23       &  18.57   \\
SC  &  23.59       &  10.93    \\
HS  &  27.83       &  19.50   \\
HS\&R  & 28.68     &  19.71   \\
\midrule
ItD-IO  &  32.49   &  20.05    \\
ItD  &  \textbf{38.70}   &    \textbf{21.60}    \\
\bottomrule
\end{tabular}
\end{adjustbox}
\caption{The main results of our experiments and the Effectiveness of Deductive Data Generation and Naive Bayesian Induction. ItD is superior to all of the previous methods on both datasets, while both Deductive Data Generation and Naive Bayesian Induction effectively contribute to the performance of ItD.}
\label{main results}
\end{table}


\begin{figure*}[t]
    \centering
    \includegraphics[width=0.9\textwidth]{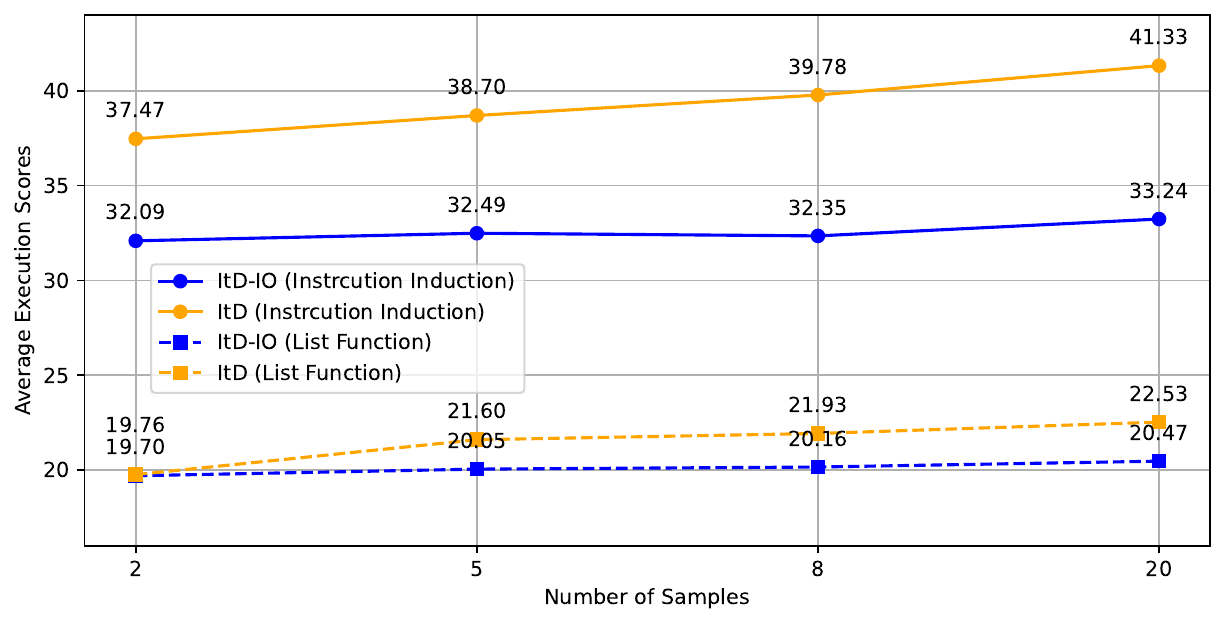}
    \caption{Naive Bayesian Induction can benefit from the increase in the number of observed samples.}
    \label{sample_number}
\end{figure*}

\begin{table*}[t]
\centering
\begin{tabular}{l|ccc|cc}
\hline
Dataset & \multicolumn{3}{c|}{Instruction Induction} & \multicolumn{2}{c}{List Function} \\
\hline
Model & Llama-2-7b-chat& Llama-2-13b-chat& ChatGPT*& Mixtral-8x7B& ChatGPT* \\
\hline
IO & 13.23 & 34.43 & 56.75 & 18.57 & 26.88 \\
ItD & 38.71 & 44.64 & 62.07 & 21.60 & 29.59 \\
\hline
Human & \multicolumn{3}{c|}{67.82} & \multicolumn{2}{c}{37.08} \\
\hline
\end{tabular}
\caption{ItD is effective for LLMs of different sizes. * denotes that models only use the ItD-IO version as we are not able to modify the decoding algorithms of these black-box LLMs. Human denotes the results that the Reasoner $\mathcal{R}$ directly adopts the human-written references for evaluation.}
\label{model size}
\end{table*}

\begin{table}[t]
\centering
\begin{adjustbox}{width=\columnwidth}
\begin{tabular}{lccc}
\toprule
Dataset        & Instruction Induction & List Function \\
\midrule
Model          & Llama-2-7b-chat    & Mixtral-8x7B \\
\midrule
HS           & 27.83     &    19.50       \\
HS+D  & 31.76     &     20.30        \\
ItD          & 38.70     &     21.60       \\
ItD+D & 41.01     &     23.91       \\
\bottomrule
\end{tabular}
\end{adjustbox}
\caption{Both ItD and baseline method HS can benefit from a more powerful deductor (ChatGPT, denoted as +D). Compared with conducting deduction by the tested model, both methods with ChatGPT helping in conducting deduction will have better performances.}
\label{deductor}
\end{table}

\subsection{Baselines}
We adopt the following baselines to compare with our proposed ItD:
\begin{itemize}[itemsep=1pt,topsep=1pt,parsep=0pt,leftmargin=*]
    \item \textbf{IO (input-output}, \citealt{honovich2022instruction}). This baseline is the plain prompt, i.e. directly splice the observations $x_1,y_1,x_2,y_2,...,x_n,y_n$ as the IO prompt, and feed this prompt for LLMs to conduct induction.
    \item \textbf{SC (self-consistency}, \citealt{self-consistency}). Based on the IO prompt, the SC method will sample $k$ hypotheses and select the most consistent one by taking a majority vote.
    \item \textbf{HS (hypothesis search}, \citealt{wang2023hypothesis}). Based on the IO prompt, the HS method will evaluate the generated hypotheses by applying the hypotheses to the observed samples. The deductive reasoning results will be used to filter out the most qualified hypothesis.  
    \item \textbf{HS\&R (hypothesis search \& refinement}, \citealt{qiu2023phenomenal}). After selecting the best hypothesis, this baseline allows LLMs to refine the hypothesis to a better one based on the execution results.
\end{itemize}



\subsection{Main Results}
\label{sota}
We first compare ItD with previous methods to see whether ItD trains the LLMs to become better inductors.

For Instruction Induction, we adopt Llama-2-7b-chat as the LLM for all methods.
For List Function,
as List Function is a task of symbolic reasoning and LLMs are found poor at symbolic deduction than semantic deduction \cite{semanticThanSymbolic}, we adopt a larger and more powerful LLM, Mixtral-8x7B, for all methods.

As shown in Table~\ref{main results}, ItD is significantly superior to all existing methods on both datasets, bringing relative performance improvement of 193\% and 16\% compared with the base model (IO), while bringing relative performance improvement of 35\% and 10\% compared with the previous SOTA (HS\&R).
These results verify that ItD is better than previous methods in empowering the inductive capability of LLMs.

\subsection{The Effectiveness of Deductive Data Generation}


To verify the effectiveness of Deductive Data Generation, we here compare the ItD-IO version with the base model (IO). The only difference between these two models is that ItD-IO is fine-tuned with the data generated by Deductive Data Generation with the IO prompt.

As shown in Table~\ref{main results}, ItD-IO is superior to the base model on both datasets, bringing the relative performance improvement of 146\% and 8\%. 
These results indicate that Deductive Data Generation can produce effective fine-tuning data for the LLMs.

\subsection{The Effectiveness of Naive Bayesian Induction}


The Naive Bayesian Induction allows us to optimize the use of each observed sample and to take advantage of the increase in the number of observed samples.
To verify the effectiveness of Naive Bayesian Induction, we first compare the complete ItD with ItD-IO.
As shown the Table~\ref{main results}, the model trained with complete ItD significantly outperforms ItD-IO on both datasets, indicating the effectiveness of Naive Bayesian Induction. 

Moreover, we conduct experiments to verify that Naive Bayesian Induction can benefit from the increase in the number of observed samples.
While ItD-IO is tuned with 5 pairs of $(x,y)$ per batch, we test both ItD-IO and ItD with 2, 5, 8, and 20 pairs of $(x,y)$ per batch.
As shown the Figure~\ref{sample_number}, the performance of ItD-IO remains almost unchanged with the increase in the number of samples, with ItD-IO-20 only outperforming ItD-IO-2 by 1.15\% and 0.77\%.
In contrast, the performance of ItD enjoys a natural improvement as the number of samples grows, with ItD-20 outperforming ItD-2 by 3.86 \% and 2.77\%.
These results verify the effectiveness of Naive Bayesian Induction in both directly improving the induction performance and making LLMs capable of taking advantage of the increase in the number of observed samples.

\subsection{Discussion}

\subsubsection{The Effectiveness of ItD on Different Sizes of LLMs}

To verify whether ItD is effective with different sizes of LLMs, we adopt extra LLMs of different sizes for each task: 
For Instruction Induction, besides Llama-2-7b-chat, we adopt Llama-2-13b-chat, and ChatGPT for experiments. 
For the List Function, besides Mixtral-8x7B, we adopt ChatGPT for the experiments of this task. 
Note that for ChatGPT, as we are not able to modify the decoding algorithm during its inference time, we only apply the ItD-IO version for it. We use the official API for the fine-tuning and inference of the ChatGPT.

As shown in Table~\ref{model size}, for LLMs of different sizes, ItD can effectively enhance the performance of the model, with the relative performance improvement ranging from 9\% to 193\% across different models. These results support that ItD can effectively empower the inductive capability of LLMs of different sizes.

\subsubsection{A More Powerful Deductor Can Bring Further Improvements for ItD}

Both HS and ItD need a deductor to improve the induction process. For HS, the deductor is used to search for the best-proposed hypothesis by evaluating them on the observed samples. For ItD, the deductor is used to deduce data for fine-tuning. In the experiments above, the deductor used in these two methods is both the tested model itself. However, here we would like to discuss whether a more powerful deductor will further improve these methods. So we adopt the Reasoner $\mathcal{R}$ of the tasks, i.e. ChatGPT, as a more powerful deductor for these methods as the comparison.

As shown in Table~\ref{deductor}, After being equipped with a more powerful deductor (denoted as +D), both HS and ItD gain performance improvements on both datasets, while ItD still consistently outperforms HS whatever the deductor is the base model or ChatGPT. These results further inform us that the more powerful the Deductor, the better it helps in training the Inductor.



\section{Related Work}

\subsection{Capability of Induction of LLMs}
Although LLMs have shown great power in a large number of fields of NLP, it is shown by previous research that they are poor on induction. 
\citealt{eval1} and \citealt{eval2} found that LLMs are poor on abstract induction tasks like Abstraction and Reasoning Corpus \cite{chollet2019measure}. 
Another research \cite{mitchell2023comparing} found that even GPT-4 and GPT-4V are still not able to robustly form abstractions and reason in contexts not previously seen in their training data.
However,
\citealt{bang2023multitask} and \citealt{semanticThanSymbolic} have made quantitative evaluations on LLMs and found that they are much better at deduction than induction.
Inspired by the findings of these works, we propose a novel framework, ItD, to leverage the powerful deductive capability of LLMs to enhance their inductive capability.

\subsection{Memory-Oriented Induction}
LLMs have shown strong ability in reasoning in various down-steam tasks. However, they still struggle when it comes to an unfamiliar task. Thus, many previous works have designed a working memory to help LLMs store and use task-specific knowledge \cite{yang2023failures, sun2023expnote, hit, zhao2023expel}. The LLMs are prompted to induce task-specific knowledge in the form of facts or rules and store them in the memory during the induction stage. In the deductive reasoning stage, a retriever will be called to retrieve relevant knowledge about the current question from the memory and prompt it to the LLMs. 
For these applications, ItD is supposed to be a powerful framework for these methods to tune the LLMs to gain better inductive capability to further improve their performances.

\subsection{Hypothesis Search and Refinement}
Some previous works have proposed methods to improve the induced hypotheses of LLMs by conducting Hypothesis Search and Refinement. Hypothesis Search \cite{wang2023hypothesis} proposes to implement the natural language hypothesis to the Python program and then execute them on the observed samples, the execution results are then used to filter out the better hypotheses. Based on Hypothesis Search, Iterative Hypothesis Refinement \cite{qiu2023phenomenal} proposes to iteratively refine the hypothesis through LLMs based on the feedback of execution results. Compared with these methods, ItD improves the inherent inductive capability of LLMs by fine-tuning them with high-quality deduced data and producing a better induction algorithm.

\subsection{Naive Bayes-based Context Extension}
NBCE \cite{nbce} is recently proposed as an effective method to extend the context for LLMs. It is proposed for the scenes of conducting QA with a batch of documents. However, the documents are likely to be coupled with others and thus cause NBCE poorly infer the answers. Compared with NBCE, Naive Bayesian Induction applies this derivation to the problem of induction, where the samples are conditionally independent of each other given $f$ in nature.
Moreover, we involve the tuning process with GD prompt in ItD, which not only 
optimize the use of each observed sample but also take advantage of the increase in the number of samples.

\section{Conclusion}
In this paper, we propose a novel framework, ItD, to enable LLMs to teach themselves induction through deduction. We conduct a series of experiments on two types of induction datasets and verify that ItD is superior to existing methods in empowering the inductive capability of LLMs. Moreover, we verify the effectiveness of Deductive Data Generation and Naive Bayesian Induction. More experiment results support that ItD can be effectively applied to LLMs of different sizes, and a more powerful deductor can further improve the performance of ItD.

\section*{Limitations}
With our ItD framework, we can improve both the symbolic deductive reasoning and semantic deductive reasoning tasks. However, constrained by the limited capability of LLMs in symbolic reasoning, the performance of ItD on List Function (a symbolic deductive task) is not as satisfying as it is on Instruction Induction (a semantic deductive task). Besides, our proposed Naive Bayesian Group Decoding is still categorized to greedy algorithms. It does not involve planning and may likely fall into local optima. We leave further exploration of these directions as future work.

\section*{Ethics Statement}
This paper proposes a method for LLMs to teach themselves induction through deduction. All experiments are conducted on publicly available datasets. Thus there is no data privacy concern. Meanwhile, this paper does not involve human annotations, and there are no related ethical concerns.



\bibliography{anthology,custom}

\bibliographystyle{acl_natbib}

\appendix

\section{Setups}
\label{setups}
For Instruction Induction, it contains 24 sub-tasks, with the induction set $\mathcal{D}_{in}$ of each sub-task including 100 batches, each batch includes $n=5$ pairs of $(x,y)$. The deduction set $\mathcal{D}_{de}$ of each sub-task including 100 pairs of $(x,y)$ for testing. 
For List Function, it contains 250 sub-tasks, with the induction set $\mathcal{D}_{in}$ of each sub-task including 3 batches, each batch includes $n=5$ pairs of $(x,y)$. The deduction set $\mathcal{D}_{de}$ of each sub-task including 17 pairs of $(x,y)$ for testing.

Induction on both tasks is conducted in a zero-shot manner by LLMs. For both ordinary beam search and Naive Bayesian Group Decoding used during the induction phase, we adopt the beam size of 5.

For Deductive Data Generation, we adopt top-p = 0.95 and temperature = 0.3 to sample 5 transformations $f$ from each batch of $\mathcal{D}_{in}$. And then we generate 5 pairs of $(x,y)$ for each $f$.
For both ItD and ItD-IO, we fine-tune them using the same data above, with a learning rate of 1e-4 and for 3 epochs.
For Naive Bayesian Group Decoding, we create a patch for the utils.py in the transformer library, it can be easily installed and uninstalled using our scripts.

The prompts used in Induction (\S\ref{NBGD}) and Deduction with In-Context Learning (\S\ref{DICL}) for Instruction Induction and List Function are shown in Table~\ref{insin_prompt} and Table~\ref{listfunc_prompt}, respectively. Note that the text in the Induction part is shared by both the IO prompt and the GD prompt (for the IO prompt, $n>1$, and for the GD prompt, $n=1$).

\begin{table}[t]
\centering
\begin{adjustbox}{width=\columnwidth}
\begin{tabular}{lcll}
\hline
Dataset   & \multicolumn{1}{c}{Instruction Induction}                                                                                                                   \\ \hline
Induction & \multicolumn{1}{l}{\begin{tabular}[c]{@{}l@{}}   

I gave a friend an instruction and an input. \\
The friend read the instruction \\ and wrote an output for the input.\\
Here is the input-output pair:\\
Input: $\{{x_1}\}$\\ Output: $\{{y_1}\}$\\ ...... \\ Input: $\{{x_n}\}$\\ Output: $\{{y_n}\}$\\
The instruction was                                                          
            \end{tabular}}                                                                                                                                \\

\hline
\begin{tabular}[c]{@{}l@{}}Deduction\\ with\\ In-Context \\Learning\end{tabular} & \multicolumn{1}{l}{\begin{tabular}[c]{@{}l@{}}                         

You are a smart assistant, \\
now please help me generate corresponding \\
input-output pairs that satisfy the given instruction. \\
Do not repeat the instructions in the inputs. \\
instruction: describe the major color of the given object. \\
Input: watermelon. \\
Output: green.\\
Input: panda.\\
Output: black and white.\\
Input: ocean.\\
Output: blue.\\
Input: blood.\\
Output: red.\\
Input: snow.\\
Output: white.\\
instruction: answer the capital of the given country.\\
Input: USA\\
Output: Washington.\\
Input: China.\\
Output: Beijing.\\
Input: Russia.\\
Output: Moscow.\\
Input: France.\\
Output: Paris.\\
Input: UK.\\
Output: London.\\
Instruction: $\{f\}$
\end{tabular}}                                                                     
\\ \hline

\end{tabular}
\end{adjustbox}
\caption{The prompts used for Instruction Induction.}
\label{insin_prompt}
\end{table}

\begin{table}[t]
\centering
\begin{adjustbox}{width=\columnwidth}
\begin{tabular}{lcll}
\hline
Dataset   & \multicolumn{1}{c}{List Function}                                                                                                                                                                      \\ \hline
Induction & \multicolumn{1}{l}{\begin{tabular}[c]{@{}l@{}}There is a transformation that transforms \\ the input list to the output list.\\ please tell me the transformation in natural language.\\ Input: $\{{x_1}\}$\\ Output: $\{{y_1}\}$\\ ...... \\ Input: $\{{x_n}\}$\\ Output: $\{{y_n}\}$\\ The transformation is:\\ The transformation\end{tabular}}                                                                                                                                                        \\

\hline

\begin{tabular}[c]{@{}l@{}}Deduction\\ with\\ In-Context \\Learning\end{tabular}   & \multicolumn{1}{l}{\begin{tabular}[c]{@{}l@{}}You are a smart assistant, \\ now please help me predict the output \\ given the input and the transformation.\\ transformation: Remove the first and the second element.\\ input: {[}0, 8, 9, 3, 7, 5, 5{]}\\ output: {[}9, 3, 7, 5, 5{]}\\ input: {[}7, 3, 9, 6{]}\\ output: {[}9, 6{]}\\ input: {[}0, 0, 0, 7, 7, 7{]}\\ output: {[}0, 7, 7, 7{]}\\ input: {[}2, 5, 5, 6, 3{]}\\ output: {[}5, 6, 3{]}\\ input: {[}7, 3, 6, 8, 8, 5, 0{]}\\ output: {[}6, 8, 8, 5, 0{]}\\ \\ transformation: Retain the elements that greater than 5.\\ input: {[}3, 4, 8, 1, 0, 5, 3, 7, 9, 9{]}\\ output: {[}8, 7, 9, 9{]}\\ input: {[}0, 4, 5, 7, 7, 1, 2, 6{]}\\ output: {[}7, 7{]}\\ input: {[}1, 0, 0, 3, 7, 8, 5{]}\\ output: {[}7, 8{]}\\ input: {[}5, 1, 9, 3, 6, 1, 7, 3{]}\\ output: {[}9, 6, 7{]}\\ input: {[}2, 6, 8, 1, 7{]}\\ output: {[}6, 8, 7{]}\\ \\ transformation: Reverse the input list.\\ input: {[}1, 0, 3, 8{]}\\ output: {[}8, 3, 0, 1{]}\\ input: {[}1, 3, 7, 4, 2, 0, 8, 9{]}\\ output: {[}9, 8, 0, 2, 4, 7, 3, 1{]}\\ input: {[}8, 9, 0, 1, 3{]}\\ output: {[}3, 1, 0, 9, 8{]}\\ input: {[}5, 5, 6, 8, 0, 1, 3, 2{]}\\ output: {[}2, 3, 1, 0, 8, 6, 5, 5{]}\\ input: {[}2, 0, 8, 7, 5, 4{]}\\ output: {[}4, 5, 7, 8, 0, 2{]}\\ \\ transformation: Append 5 to the input list.\\ input: {[}7, 0, 3, 6{]}\\ output: {[}7, 0, 3, 6, 5{]}\\ input: {[}1, 2, 3, 7, 8, 5{]}\\ output: {[}1, 2, 3, 7, 8, 5, 5{]}\\ input: {[}2, 9, 6, 3, 7, 5, 4, 4{]}\\ output: {[}2, 9, 6, 3, 7, 5, 4, 4, 5{]}\\ input: {[}0, 0, 8, 6, 9{]}\\ output: {[}0, 0, 8, 6, 9, 5{]}\\ input: {[}7, 5, 6, 5, 3, 3, 2{]}\\ output: {[}7, 5, 6, 5, 3, 3, 2, 5{]} \\ \\transformation: $\{f\}$\end{tabular}} \\ \hline
\end{tabular}
\end{adjustbox}
\caption{The prompts used for List Function.}
\label{listfunc_prompt}
\end{table}

\section{Detailed Results}
\label{detailed results}

The detailed results of Instruction Induction are shown in Table~\ref{detail_insin}. As the List Function contains 250 sub-tasks and we have 19 methods in all, the table of its detailed results will be too large for the paper. Instead, you can find it at https://anonymous.4open.science/r/ItD-E844.

\begin{table*}[t]
\centering
\scriptsize
    \begin{tabular}{lccccccc}
    \hline
        Task & IO (L7) & SC (L7) & HS (L7) & HS\&R (L7) & HS+D (L7) & ItD (L7) & ItD+D (L7) \\ \hline
        active\_to\_passive & 56.18 & 3.23 & 9.55 & 20.1 & 16.13 & 90.15 & 100.00 \\ \hline
        antonyms & 40.82 & 79.75 & 81.59 & 80.97 & 83.50 & 80.40 & 83.00 \\ \hline
        cause\_and\_effect & 16.74 & 15.70 & 24.78 & 21.52 & 28.78 & 45.84 & 57.06 \\ \hline
        common\_concept & 0.96 & 6.47 & 7.00 & 6.98 & 6.67 & 17.58 & 3.21 \\ \hline
        diff & 1.14 & 3.32 & 9.78 & 15.46 & 14.70 & 17.00 & 34.49 \\ \hline
        first\_word\_letter & 12.56 & 58.03 & 58.90 & 54.28 & 81.49 & 71.27 & 100.00 \\ \hline
        informal\_to\_formal & 40.51 & 34.59 & 40.99 & 43.16 & 43.04 & 26.22 & 48.16 \\ \hline
        larger\_animal & 0.04 & 11.86 & 22.34 & 23.29 & 28.54 & 6.05 & 30.06 \\ \hline
        letters\_list & 0.12 & 0.31 & 1.15 & 1.23 & 1.22 & 0.04 & 0.00 \\ \hline
        negation & 9.52 & 6.51 & 13.16 & 14.44 & 15.39 & 44.95 & 43.45 \\ \hline
        num\_to\_verbal & 3.00 & 3.00 & 4.00 & 7.58 & 7.48 & 97.00 & 100.00 \\ \hline
        orthography\_starts\_with & 3.02 & 6.94 & 5.71 & 7.26 & 9.76 & 1.86 & 43.20 \\ \hline
        rhymes & 26.64 & 2.93 & 3.00 & 2.72 & 2.45 & 0.19 & 2.25 \\ \hline
        second\_word\_letter & 4.73 & 2.13 & 1.23 & 1.92 & 5.36 & 8.64 & 2.28 \\ \hline
        sentence\_similarity & 0.00 & 0.00 & 0.00 & 0.00 & 0.00 & 0.00 & 0.00 \\ \hline
        sentiment & 2.01 & 17.49 & 26.41 & 27.18 & 45.82 & 87.49 & 36.96 \\ \hline
        singular\_to\_plural & 28.58 & 94.05 & 97.93 & 96.94 & 98.78 & 97.86 & 99.95 \\ \hline
        sum & 22.33 & 50.54 & 83.64 & 84.37 & 84.13 & 50.02 & 10.37 \\ \hline
        synonyms & 8.56 & 0.39 & 1.22 & 1.05 & 1.43 & 2.79 & 1.42 \\ \hline
        taxonomy\_animal & 0.00 & 0.37 & 1.56 & 0.77 & 2.27 & 1.59 & 8.08 \\ \hline
        translation\_en-de & 11.50 & 48.37 & 50.42 & 51.97 & 52.33 & 50.82 & 53.78 \\ \hline
        translation\_en-es & 14.99 & 65.50 & 67.83 & 66.41 & 69.15 & 57.50 & 60.08 \\ \hline
        translation\_en-fr & 13.52 & 47.99 & 47.68 & 51.86 & 52.45 & 35.96 & 46.37 \\ \hline
        word\_in\_context & 0.00 & 6.74 & 8.08 & 6.80 & 11.30 & 37.61 & 20.02 \\ \hline
        average & 13.22782 & 23.59205 & 27.83108 & 28.67735 & 31.75728 & 38.70161 & 41.00783 \\ \hline
        Task & ItD-IO-2 (L7) & ItD-IO-5 (L7) & ItD-IO-8 (L7) & ItD-IO-20 (L7) & ItD-2 (L7) & ItD-8 (L7) & ItD-20 (L7) \\ \hline
        active\_to\_passive & 37.11 & 56.76 & 56.58 & 53.34 & 79.82 & 93.25 & 98.98 \\ \hline
        antonyms & 77.99 & 67.24 & 66.55 & 75.45 & 76.79 & 82.48 & 83.29 \\ \hline
        cause\_and\_effect & 26.96 & 22.44 & 18.28 & 12.44 & 42.14 & 42.10 & 36.98 \\ \hline
        common\_concept & 4.45 & 5.02 & 4.87 & 7.66 & 16.69 & 17.72 & 17.72 \\ \hline
        diff & 0.93 & 2.17 & 8.74 & 5.94 & 19.00 & 16.00 & 37.00 \\ \hline
        first\_word\_letter & 51.96 & 43.84 & 33.36 & 26.69 & 65.93 & 79.49 & 69.21 \\ \hline
        informal\_to\_formal & 35.36 & 31.87 & 33.31 & 38.23 & 27.13 & 26.03 & 23.73 \\ \hline
        larger\_animal & 41.31 & 20.86 & 12.21 & 16.35 & 8.54 & 3.39 & 0.00 \\ \hline
        letters\_list & 0.00 & 0.00 & 0.00 & 0.22 & 0.04 & 0.06 & 0.07 \\ \hline
        negation & 33.47 & 31.54 & 39.46 & 29.33 & 50.86 & 50.45 & 56.17 \\ \hline
        num\_to\_verbal & 98.30 & 96.86 & 96.63 & 99.01 & 96.00 & 99.00 & 100.00 \\ \hline
        orthography\_starts\_with & 7.42 & 3.61 & 2.87 & 2.94 & 2.51 & 2.12 & 1.01 \\ \hline
        rhymes & 1.64 & 0.75 & 0.67 & 0.87 & 0.43 & 0.20 & 0.04 \\ \hline
        second\_word\_letter & 3.55 & 4.25 & 1.43 & 0.00 & 8.94 & 8.79 & 12.54 \\ \hline
        sentence\_similarity & 0.05 & 0.00 & 0.02 & 0.00 & 0.00 & 0.00 & 0.00 \\ \hline
        sentiment & 21.75 & 31.59 & 35.10 & 61.95 & 80.33 & 88.11 & 89.00 \\ \hline
        singular\_to\_plural & 91.11 & 97.46 & 96.77 & 91.52 & 97.39 & 97.37 & 97.35 \\ \hline
        sum & 68.16 & 64.07 & 67.33 & 85.26 & 42.02 & 52.00 & 75.00 \\ \hline
        synonyms & 4.92 & 8.32 & 6.59 & 6.81 & 2.01 & 3.05 & 2.09 \\ \hline
        taxonomy\_animal & 6.21 & 2.77 & 0.96 & 0.13 & 2.16 & 1.11 & 0.46 \\ \hline
        translation\_en-de & 50.54 & 56.46 & 55.35 & 55.65 & 51.67 & 52.77 & 51.39 \\ \hline
        translation\_en-es & 55.85 & 59.43 & 60.74 & 51.04 & 58.11 & 58.79 & 58.02 \\ \hline
        translation\_en-fr & 40.11 & 48.32 & 52.40 & 52.22 & 35.04 & 37.84 & 35.81 \\ \hline
        word\_in\_context & 11.11 & 24.01 & 26.22 & 24.85 & 35.82 & 42.67 & 46.14 \\ \hline
        average & 32.09425 & 32.48515 & 32.35150 & 33.24609 & 37.47368 & 39.78301 & 41.33348 \\ \hline
        Task & IO (L13) & ItD (L13) & IO (ChatGPT) & ItD-IO (ChatGPT) & IO (reference) & ~ & ~ \\ \hline
        active\_to\_passive & 93.43 & 100.00 & 100.00 & 100.00 & 100.00 & ~ & ~ \\ \hline
        antonyms & 73.36 & 81.23 & 77.54 & 73.80 & 81.11 & ~ & ~ \\ \hline
        cause\_and\_effect & 10.52 & 57.16 & 30.24 & 44.04 & 39.33 & ~ & ~ \\ \hline
        common\_concept & 3.91 & 9.17 & 7.84 & 9.21 & 12.00 & ~ & ~ \\ \hline
        diff & 32.57 & 91.56 & 93.00 & 99.00 & 99.89 & ~ & ~ \\ \hline
        first\_word\_letter & 26.10 & 9.13 & 100.00 & 100.00 & 99.89 & ~ & ~ \\ \hline
        informal\_to\_formal & 52.01 & 42.38 & 54.56 & 53.94 & 59.55 & ~ & ~ \\ \hline
        larger\_animal & 35.98 & 87.67 & 68.95 & 77.73 & 91.78 & ~ & ~ \\ \hline
        letters\_list & 3.59 & 7.03 & 77.88 & 94.02 & 89.44 & ~ & ~ \\ \hline
        negation & 52.05 & 61.44 & 75.09 & 73.03 & 74.50 & ~ & ~ \\ \hline
        num\_to\_verbal & 44.18 & 100.00 & 99.90 & 100.00 & 93.00 & ~ & ~ \\ \hline
        orthography\_starts\_with & 2.20 & 12.12 & 25.28 & 40.42 & 52.50 & ~ & ~ \\ \hline
        rhymes & 1.17 & 0.18 & 1.75 & 6.67 & 11.38 & ~ & ~ \\ \hline
        second\_word\_letter & 1.08 & 0.12 & 50.60 & 85.81 & 99.00 & ~ & ~ \\ \hline
        sentence\_similarity & 0.00 & 0.00 & 0.00 & 0.00 & 0.33 & ~ & ~ \\ \hline
        sentiment & 74.84 & 39.01 & 53.82 & 66.81 & 82.75 & ~ & ~ \\ \hline
        singular\_to\_plural & 82.45 & 100.00 & 94.74 & 94.53 & 99.88 & ~ & ~ \\ \hline
        sum & 7.81 & 20.74 & 97.00 & 100.00 & 98.87 & ~ & ~ \\ \hline
        synonyms & 3.95 & 5.56 & 14.72 & 15.28 & 12.88 & ~ & ~ \\ \hline
        taxonomy\_animal & 0.56 & 0.35 & 37.99 & 52.72 & 94.00 & ~ & ~ \\ \hline
        translation\_en-de & 58.52 & 60.77 & 62.12 & 62.66 & 61.83 & ~ & ~ \\ \hline
        translation\_en-es & 61.95 & 73.97 & 73.61 & 74.33 & 73.50 & ~ & ~ \\ \hline
        translation\_en-fr & 57.87 & 66.63 & 64.46 & 65.51 & 69.25 & ~ & ~ \\ \hline
        word\_in\_context & 46.31 & 45.04 & 0.97 & 0.28 & 30.90 & ~ & ~ \\ \hline
        average & 34.43337 & 44.63556 & 56.75255 & 62.07481 & 67.81509 \\ \hline
    \end{tabular}
\caption{Detailed results of Instruction Induction. L7 denotes Llama-2-7b-chat and L13 denotes Llama-2-13b-chat.}
\label{detail_insin}
\end{table*}



\end{document}